\pdfoutput=1   
\documentclass[11pt,a4paper]{article}

\usepackage[utf8]{inputenc}
\usepackage[T1]{fontenc}
\usepackage{newtxtext}

\usepackage[top=1.35cm,bottom=2.5cm,left=2.4cm,right=2.4cm]{geometry}

\usepackage{amsmath,amsfonts,amssymb,amsthm}

\usepackage{graphicx}
\usepackage{float}
\usepackage{booktabs}
\usepackage{tabularx}
\usepackage{array}

\usepackage[dvipsnames]{xcolor}
\usepackage[colorlinks=true,linkcolor=blue,citecolor=blue,urlcolor=blue]{hyperref}

\usepackage[numbers,sort&compress]{natbib}

\usepackage{authblk}

\usepackage{microtype}
\usepackage{url}

\newcommand{\benchname}{CrystalXRD-Bench}
\newcommand{\papertitleplain}{\benchname{}: Benchmarking Vision-Language Models for XRD Peak Indexing Across Diverse Crystalline Materials}
\newcommand{\papertitledisplay}{\benchname{}: Benchmarking Vision-Language Models\\for XRD Peak Indexing Across Diverse Crystalline Materials}

\title{%
\texorpdfstring{%
\noindent\makebox[\textwidth]{%
\raisebox{-0.06cm}{\includegraphics[height=0.95cm]{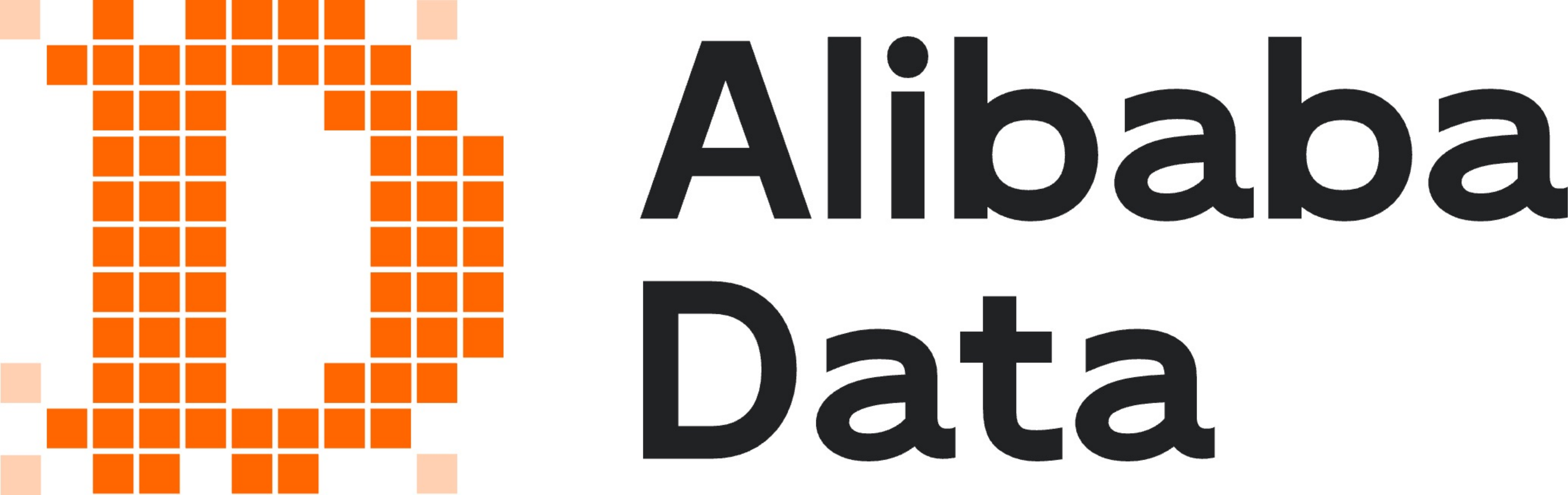}}%
\hfill
\raisebox{-0.03cm}{\includegraphics[height=1.45cm]{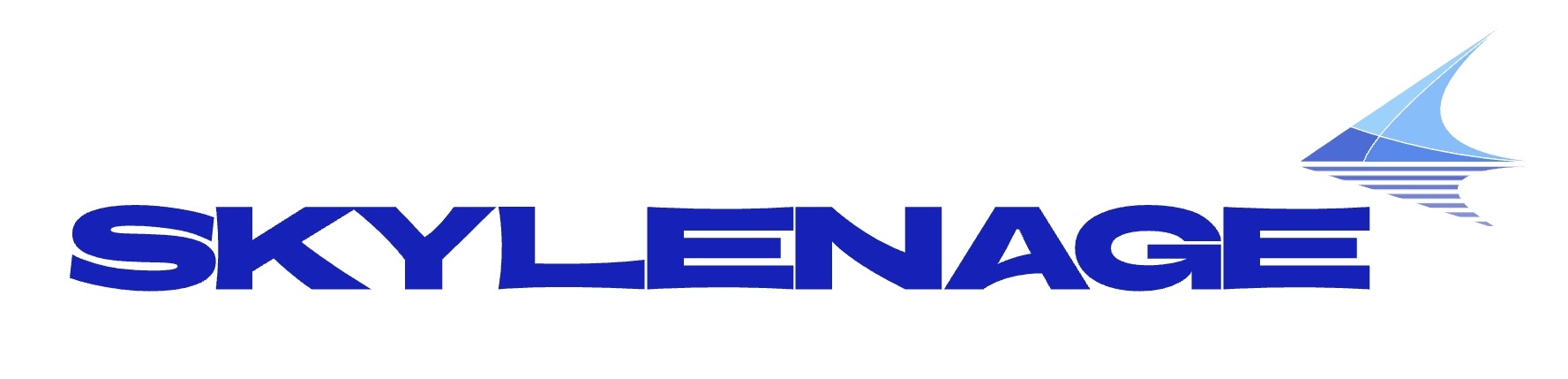}}%
}\\[-0.2em]
\rule{\textwidth}{0.5pt}\\[0.7em]
\papertitledisplay
}{\papertitleplain}%
}

\author[1,$\dagger$]{Chengliang Xu}
\author[1,$\dagger$]{Xiaogang Li}
\author[1]{Peiyao Xiao}
\author[1]{Beng Wang}
\author[1,$*$]{Hu Wei}
\author[1,$*$]{Bing Zhao}

\affil[1]{Alibaba Group, Beijing, China}

\date{%
\vspace{0.3em}
{\small $^\dagger$Equal contribution \quad $^*$Corresponding authors}\\[0.3ex]
{\footnotesize \texttt{clxu@nao.cas.cn}, \texttt{lixiaogang.lxg@alibaba-inc.com}, \texttt{xiaopeiyao.xpy@alibaba-inc.com},}\\[0.1ex]
{\footnotesize \texttt{yuanjian.wb@alibaba-inc.com}, \texttt{kongwang@alibaba-inc.com}, \texttt{xiongdao@alibaba-inc.com}}\\[0.3ex]
{\small Correspondence: \texttt{kongwang@alibaba-inc.com}, \texttt{xiongdao@alibaba-inc.com}}%
}

\makeatletter
\renewcommand{\maketitle}{%
  \vspace*{-1.25cm}
  \begin{center}
    {\huge\bfseries \@title \par}
    \vspace{0.9em}
    {\lineskip .5em \@author \par}
    \vspace{0.2em}
    {\@date \par}
  \end{center}
  \vspace{0.25em}
}
\makeatother

\begin{document}
\maketitle

\begin{abstract}
Miller-index identification from powder XRD patterns requires capabilities untested by existing multimodal benchmarks: the model must read a narrow peak location from a rendered scientific curve and then connect that observation to multi-step crystallographic reasoning.
We introduce \benchname{}, a 250-sample benchmark built from 10 public crystallographic databases for a single task: recover the full set of HKLs contributing to the highest-intensity peak in an XRD pattern.
Each sample pairs the rendered XRD image with the source CIF text and chemical formula, so visual extraction errors and reasoning errors can be examined side by side.
We evaluate seven vision-language models.
The best Jaccard score is 0.5888 (GPT-5.4) with an exact-match rate of 37.6\%, yet six of seven models remain below Jaccard~0.50; the task is far from solved.
Error patterns vary systematically: double-peak cases are especially brittle, recall-heavy models gain coverage by over-predicting HKLs, and access to CIF text does not close the gap in crystallographic calculation.
Alongside model rankings, the benchmark identifies the conditions under which current VLMs fail on quantitative scientific figures.
All data and evaluation code will be publicly available.\footnote{Dataset, evaluation code, and XRD patterns will be released at a permanent repository.}
\end{abstract}

\section{Introduction}
\label{sec:intro}

Powder X-ray diffraction (XRD) is one of the most widely used probes in modern materials science, supplying the geometric ground truth on which crystal-structure determination, phase identification, and quantitative compositional analysis ultimately rest. At the heart of every such analysis lies a deceptively simple operation: \emph{peak indexing} -- deciding which Miller indices $(h,k,l)$ correspond to an observed reflection at a given $2\theta$. The operation is deceptive because the precision it demands is unforgiving. In low-symmetry or high-angle systems, neighboring reflections can be separated by only a few tenths of a degree, so a $0.1^\circ$ reading error is sufficient to flip an assignment from one plane family to another, propagate into incorrect $d$-spacings, and ultimately corrupt the inferred unit cell~\cite{cullity2001xrd}. The visual challenge is further compounded by overlap: the strongest peak in a typical pattern is frequently a superposition of several closely spaced reflections, so the correct answer is rarely a single triplet but a small set whose cardinality must itself be inferred from the curve. Recovering that set therefore requires sub-degree chart reading coupled with multi-step crystallographic reasoning over Bragg's law, symmetry constraints, unit-cell geometry, and structure factors -- a combination that makes peak indexing a uniquely demanding probe of AI-assisted materials characterization, and one that few existing evaluation suites are positioned to assess.

Despite rapid progress on chart understanding by vision-language models (VLMs), no existing benchmark exercises this regime. Generic chart-oriented multimodal datasets---DVQA~\cite{kafle2018dvqa}, PlotQA~\cite{methani2020plotqa}, ChartQA~\cite{masry2022chartqa}, ChartBench~\cite{xu2023chartbench}, FigureQA~\cite{kahou2017figureqa}, MMC~\cite{liu2023mmc}, and ChartX~\cite{xia2024chartx}---pose visual question answering over plots, yet the answers they reward tolerate reading errors that would be catastrophic for crystallographic indexing; an extracted axis value that is merely ``close'' is acceptable for trend interpretation but useless when a tenth of a degree decides the assignment. Materials-oriented benchmarks have moved in a complementary but equally insufficient direction. LLM4Mat-Bench~\cite{rubungo2024llm4mat} and MatSci-NLP~\cite{song2023matscinlp} probe scientific knowledge from textual or structured numerical inputs, treating the underlying figures as already digested rather than as evaluation targets in their own right. The result is a structural blind spot in current evaluation practice: tasks that demand sub-degree visual extraction from a rendered scientific curve \emph{and} chained crystallographic reasoning over its features sit at the intersection of two evaluation traditions but are covered by neither, leaving us without a quantitative account of how today's frontier VLMs perform when both demands are imposed simultaneously on a single input.

This limitation also extends to prior machine-learning work on XRD itself. Park et al.~\cite{park2017classification} and Lee et al.~\cite{lee2020deep} classify crystal structures or identify phases directly from numerical diffraction arrays -- a fundamentally different input modality from the rendered figure that an experimentalist actually inspects, and one that bypasses the visual extraction step entirely. To bridge both gaps simultaneously, we adopt a paradigm that intentionally departs from these numerical-array pipelines: the model receives a rendered XRD plot exactly as a human practitioner would see it, paired with the source CIF text and chemical formula, and must output the full set of Miller indices contributing to the highest-intensity peak. This formulation preserves the visual modality that defines the practical task, while the accompanying CIF transforms the benchmark into a reasoning probe -- because the indices are deterministically derivable from the CIF via Bragg's law, any error reflects a failure either to read the plot at sufficient precision or to execute the crystallographic calculation chain, not a lack of structural information. Casting peak indexing as set prediction further sharpens the diagnosis: since the dominant peak is often a superposition, the model must commit not only to which planes contribute but to how many of them do, and our set-based metrics make both kinds of error visible in a single score.

Building on this formulation, we argue that XRD peak indexing is an unusually informative testbed for general-purpose VLMs, both as a scientific question and as a measurement instrument for current model capabilities. Scientifically, the task isolates a competence -- combining quantitative figure reading with multi-step physical reasoning -- that recurs across nearly every domain in which models are asked to interpret experimental measurements, from spectroscopy and chromatography to scattering and microscopy; progress here is therefore unlikely to remain confined to crystallography. Practically, automated peak indexing is the gateway operation for downstream procedures such as Rietveld refinement, phase quantification, and high-throughput screening of newly synthesized compounds, all of which become tractable only once the indices are correctly assigned; a model that solves the task end-to-end would meaningfully shorten the loop between materials synthesis and structural characterization rather than merely answer plot questions. Conversely, a benchmark that exposes \emph{where} today's VLMs break down -- at sub-degree reading, at overlap resolution, or at executing the cell-parameter-to-HKL chain -- supplies architectural and training signals that aggregate scores on coarser tasks cannot. \benchname{} is built to provide that signal: across seven frontier VLMs, the best Jaccard score is 0.5888 (GPT-5.4) and all but one model fall below 0.50, leaving the deterministic ceiling of 1.0 implied by the paired CIF largely untouched and isolating systematic failure modes that we analyze in the remainder of the paper.

\paragraph{Contributions.} Concretely, this work contributes:
\begin{itemize}
  \item \textbf{\benchname{}: 250 image-text multimodal XRD samples} drawn from 10 public crystallographic databases spanning MOFs, organic crystals, high-entropy alloys, and inorganic compounds, with diversity maintained through KMeans-based sampling.
  \item \textbf{Set-based metrics with an over-prediction penalty.} Dominant-peak indexing is cast as set prediction; we report Jaccard, Precision, Recall, and F1, each paired with a penalized variant.
  \item \textbf{Evaluation of seven frontier VLMs.} GPT-5.4, Gemini~3.1~Pro, Gemini~3.0~Pro, Qwen3.6-Plus, Doubao-Seed-2.0, Kimi-K2.5, and Claude~Opus~4.6 are tested under identical prompts and CIF inputs.
  \item \textbf{Failure-mode taxonomy.} We identify a difficult double-peak regime, systematic differences in calibration strategy, and a persistent gap between CIF access and correct crystallographic reasoning.
\end{itemize}
The remainder of the paper is organized as follows. Section~\ref{sec:benchmark} formalizes the task, details the data-construction pipeline, and defines our evaluation metrics; Section~\ref{sec:experiments} describes the seven VLMs and the inference protocol; Section~\ref{sec:results} reports overall and stratified results; Section~\ref{sec:discussion} discusses failure modes and limitations; and Section~\ref{sec:conclusion} concludes.

\section{\benchname{}}
\label{sec:benchmark}

\subsection{Task Formulation}
\label{sec:task}

We cast XRD peak indexing\footnote{In crystallography, ``indexing'' conventionally refers to determining unit-cell parameters from peak positions. We use the term more narrowly to mean identifying the Miller indices (HKL) that contribute to a given peak, following the usage in recent VLM evaluation literature.} as a set-prediction problem.
Given an input triplet $(\mathcal{I}, \mathcal{C}, f)$, where $\mathcal{I}$ is the XRD pattern image, $\mathcal{C}$ is the crystallographic information file (CIF) text, and $f$ is the chemical formula, the model must output a set of Miller indices $\hat{S} = \{(h_i, k_i, l_i)\}_{i=1}^{|\hat{S}|}$ corresponding to the crystallographic planes that contribute to the highest-intensity peak.

We focus on the highest peak because it is the most visually salient feature yet often reflects the superposition of several reflections, keeping ground-truth construction tractable without trivializing the task.
Extending the task to all peaks would multiply the annotation complexity and introduce partial-match ambiguities; restricting it to a single, well-defined peak keeps ground-truth construction unambiguous while still probing the model's ability to resolve overlapping reflections.

\subsection{Data Construction Pipeline}
\label{sec:data}

\benchname{} draws from ten publicly available crystallographic databases to ensure broad coverage across material families and structural motifs (Table~\ref{tab:datasets}): two metal-organic framework (MOF) resources (QMOF~\cite{rosen2021qmof}, hMOF~\cite{wilmer2012hmof}), one organic materials database (OMDB~\cite{borysov2017omdb}), one high-entropy alloy collection (Cantor-HEA~\cite{cheon2024cantorhea}), and six inorganic repositories (JARVIS-DFT, JARVIS-QETB~\cite{choudhary2020jarvis}, Materials Project~\cite{jain2013mp}, SNUMat~\cite{lee2023snumat}, OQMD~\cite{saal2013oqmd}, and GNoME~\cite{merchant2023gnome}). This multi-source design was motivated by the need for crystallographic heterogeneity: MOF structures tend to produce strong low-angle reflections with sparse overlap, high-entropy alloys concentrate their dominant peaks at mid-to-high $2\theta$ due to compressed unit cells, and the six inorganic repositories together span all seven crystal systems and hundreds of distinct space groups. By sampling across these regimes, the benchmark probes VLM performance under structurally diverse conditions rather than within any single material class, ensuring that aggregate scores reflect genuine cross-domain robustness rather than proficiency on a narrow distribution.

For each material, we generate a theoretical XRD pattern starting from the CIF structure via \texttt{pymatgen}~\cite{ong2013pymatgen}. The \texttt{XRDCalculator} computes diffraction peak positions and intensities for Cu~K$\alpha$ radiation ($\lambda_{\text{K}\alpha_1} = 1.54056$~\AA, $\lambda_{\text{K}\alpha_2} = 1.54439$~\AA, K$\alpha_2$/K$\alpha_1$ ratio $= 0.5$); the discrete peaks are then broadened into a continuous pattern using pseudo-Voigt profiles~\cite{thompson1987voigt} (FWHM $= 0.15^\circ$, mixing parameter $\eta = 0.4$) over $2\theta \in [2^\circ, 90^\circ]$ with a step size of $0.01^\circ$. The FWHM value lies within the range typical of well-crystallised laboratory samples and provides sufficient peak separation without introducing unrealistic sharpness; the pattern generation uses precise wavelength values from \texttt{pymatgen} defaults, while the FWHM and tolerance window are set conservatively to focus the evaluation on HKL identification rather than on peak-position precision. The resulting intensity curve is rendered as a standardised PNG with labeled axes (Intensity vs.\ $2\theta$), which constitutes the sole visual input to each VLM. We adopted this rendered-image format rather than numerical arrays deliberately, so that the visual extraction step---and any errors arising within it---remains a measurable component of model performance.

The ground truth (GT) for each sample is the \emph{union} of all Miller indices from theoretical peaks that contribute to the highest observed peak, constructed via the following procedure:
\begin{enumerate}
  \item Locate the global maximum $\theta^* = \arg\max_\theta I(\theta)$ on the synthetic curve.
  \item Collect all theoretical peaks within a tolerance window $|\theta_{\text{peak}} - \theta^*| < \delta$, where $\delta = 0.30^\circ \approx 2 \times \text{FWHM}$.
  This uniform tolerance is a deliberate simplification; an adaptive window scaled to local peak width would better capture overlap at high angles, where Caglioti broadening increases FWHM.
  \item The GT set is $S^* = \bigcup_{k \in \mathcal{W}} \text{HKL}(k)$, where $\mathcal{W}$ denotes peaks within the window. The unphysical reflection $[0\,0\,0]$ is filtered.
\end{enumerate}
This union-based construction reflects the physical reality that the highest observed peak in a diffraction pattern frequently arises from the superposition of several structurally distinct reflections falling within the instrument's resolution limit. The resulting \emph{union size} $|S^*|$ ranges from 1 to 6+, directly reflecting task difficulty: a single contributing peak is comparatively easy, whereas larger unions require the model to recognise and enumerate the subtle overlap among nearby reflections.

To select the final 250 samples from approximately 1,936 candidate materials, we apply KMeans clustering in a six-dimensional feature space---peak position ($\theta^*$), peak intensity, number of theoretical peaks, union size, crystal system (encoded), and space group number---fitting $k{=}25$ clusters within each database and retaining the sample nearest to each centroid. The goal of this procedure is coverage rather than cluster recovery: by distributing selections across the full crystallographic feature space, we guard against the benchmark being dominated by the most common structure types in any single repository. As a sanity check, silhouette scores were computed for each database-specific partition; the mean score across the ten databases is 0.228 (range: 0.201--0.297), which is moderate but acceptable given the heterogeneity of the feature space (Appendix~\ref{app:silhouette} reports per-database values). The resulting dataset contains 86 single-peak ($|S^*|{=}1$), 79 double-peak ($|S^*|{=}2$), and 85 triple-or-more-peak ($|S^*|{\geq}3$) samples, with $2\theta$ values spanning $2.0^\circ$--$61.1^\circ$ (mean $26.15^\circ$).

\subsection{Dataset Statistics}
\label{sec:stats}

\begin{table*}[t]
  \centering
  \caption{Summary of the 10 databases in \benchname{}. Each contributes 25 samples selected via KMeans clustering. The $2\theta$ range indicates the span of highest-peak positions across samples; values reflect two-decimal-place precision in the underlying data (e.g., OQMD: 11.25--42.89$^\circ$; GNoME: 11.29--42.91$^\circ$) and appear similar after rounding for databases with overlapping angular distributions. Difficulty distribution reflects union sizes: Single ($|S^*|{=}1$), Double ($|S^*|{=}2$), Triple+ ($|S^*|{\geq}3$).}
  \label{tab:datasets}
  \resizebox{\textwidth}{!}{%
  \begin{tabular}{llccccc}
    \toprule
    \textbf{Database} & \textbf{Material Type} & \textbf{Pool} & \textbf{$n$} & \textbf{$2\theta$ Range ($^\circ$)} & \textbf{S / D / T+} & \textbf{Source} \\
    \midrule
    QMOF       & Metal-organic framework  & 199 & 25 & 2.0--27.9   & \phantom{0}9 / \phantom{0}8 / \phantom{0}8 & \cite{rosen2021qmof} \\
    hMOF       & Metal-organic framework  & 199 & 25 & 3.2--17.3   & 12 / \phantom{0}8 / \phantom{0}5 & \cite{wilmer2012hmof} \\
    OMDB       & Organic crystals         & 201 & 25 & 5.4--32.5   & \phantom{0}7 / \phantom{0}9 / \phantom{0}9 & \cite{borysov2017omdb} \\
    Cantor-HEA & High-entropy alloys      & 195 & 25 & 39.8--50.0  & 10 / \phantom{0}6 / \phantom{0}9 & \cite{cheon2024cantorhea} \\
    JARVIS-DFT & Inorganic (DFT)          & 172 & 25 & 13.7--43.4  & \phantom{0}8 / \phantom{0}9 / \phantom{0}8 & \cite{choudhary2020jarvis} \\
    JARVIS-QETB& Inorganic (QE+TB)        & 200 & 25 & 12.0--61.1  & \phantom{0}9 / \phantom{0}7 / \phantom{0}9 & \cite{choudhary2020jarvis} \\
    MP         & Inorganic (general)      & 199 & 25 & 2.2--43.6   & \phantom{0}8 / \phantom{0}9 / \phantom{0}8 & \cite{jain2013mp} \\
    SNUMat     & Inorganic (HSE)          & 188 & 25 & 11.5--49.6  & \phantom{0}6 / \phantom{0}8 / 11 & \cite{lee2023snumat} \\
    OQMD       & Inorganic (quantum)      & 200 & 25 & 11.3--42.9  & \phantom{0}8 / \phantom{0}8 / \phantom{0}9 & \cite{saal2013oqmd} \\
    GNoME      & Inorganic (AI-discovered)& 183 & 25 & 11.3--42.9  & \phantom{0}9 / \phantom{0}7 / \phantom{0}9 & \cite{merchant2023gnome} \\
    \midrule
    \textbf{Total} & & \textbf{1,936} & \textbf{250} & \textbf{2.0--61.1} & \textbf{86 / 79 / 85} & \\
    \bottomrule
  \end{tabular}%
  }
\end{table*}

\begin{figure*}[t]
  \centering
  \includegraphics[width=\linewidth]{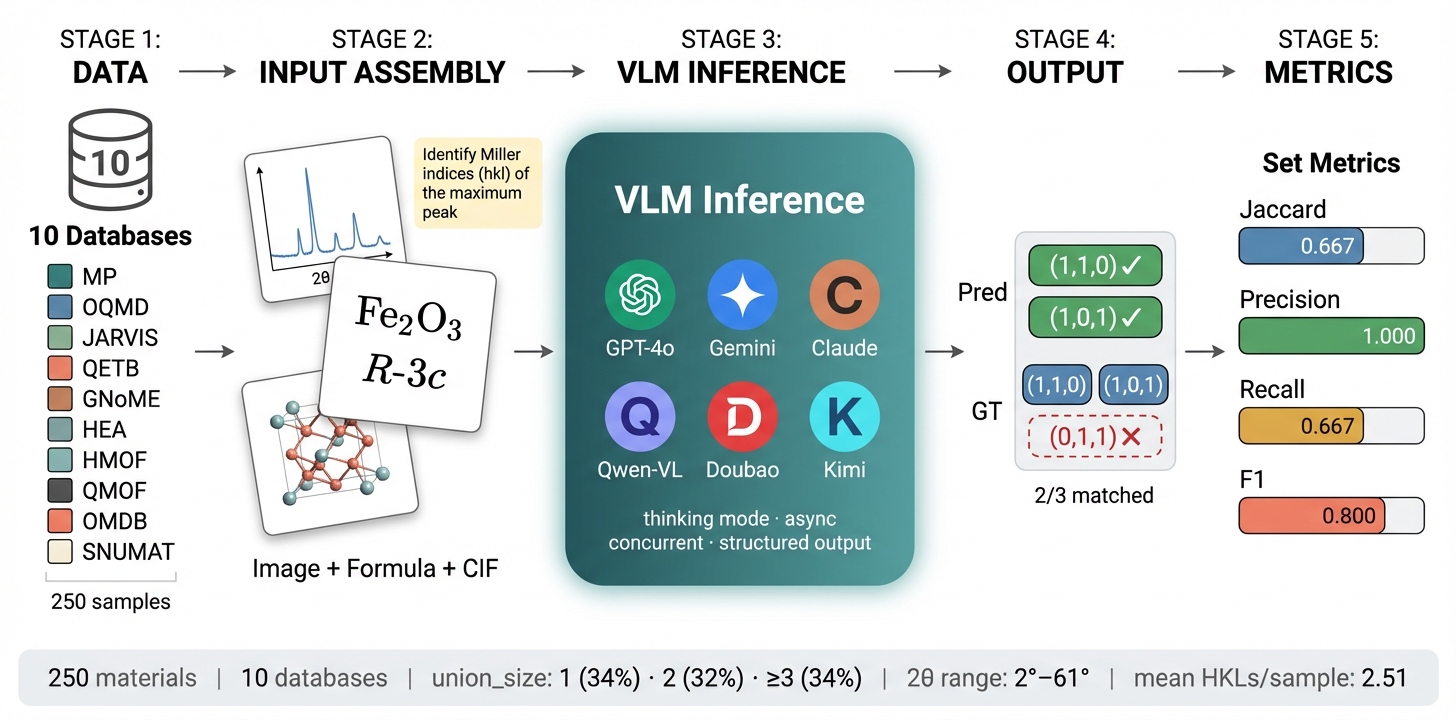}
  \caption{Pipeline of \benchname{}. Starting from CIF crystal structures from 10 databases, we generate standardized XRD patterns with pymatgen, construct HKL-union ground truth, and evaluate model outputs with set-matching metrics and over-prediction penalties.}
  \label{fig:pipeline}
\end{figure*}

\begin{figure*}[t]
  \centering
  \includegraphics[width=0.95\linewidth]{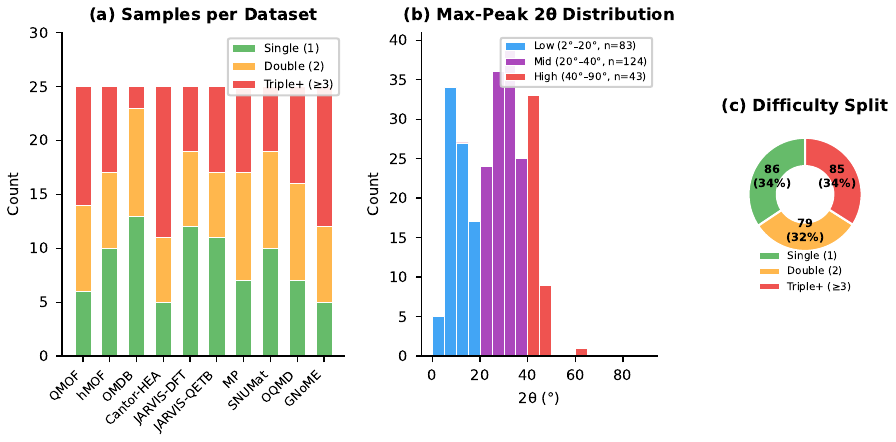}
  \caption{Dataset composition of \benchname{}: (a)~difficulty distribution (union size) per database, (b)~$2\theta$ distribution of highest-peak positions, (c)~overall difficulty breakdown (Single: 34.4\%, Double: 31.6\%, Triple+: 34.0\%).}
  \label{fig:dataset_stats}
\end{figure*}

Figure~\ref{fig:pipeline} illustrates the end-to-end \benchname{} pipeline, and Figure~\ref{fig:dataset_stats} summarizes the dataset composition.
The benchmark includes both 3-index Miller notation ($hkl$; 224 samples) and 4-index Miller-Bravais notation ($hkil$; 26 samples from hexagonal/trigonal systems).
The average GT set size is 2.51 HKL per sample.
Across peak positions, 83 samples (33.2\%) fall in the low-angle range ($2^\circ$--$20^\circ$), 124 (49.6\%) in the mid-angle range ($20^\circ$--$40^\circ$), and 43 (17.2\%) in the high-angle range ($40^\circ$--$90^\circ$).

\subsection{Evaluation Metrics}
\label{sec:metrics}

We evaluate HKL identification as set matching between the predicted set $P$ and ground-truth set $G$, after deduplication and filtering of $[0\,0\,0]$. Let $T = P \cap G$ and $U = P \cup G$.

\paragraph{Base Metrics.}
\begin{align}
  \text{Jaccard}(P,G) &= \tfrac{|T|}{|U|}, &
  \text{Precision} &= \tfrac{|T|}{|P|}, \notag \\
  \text{Recall} &= \tfrac{|T|}{|G|}, &
  \text{F1} &= \tfrac{2\,\text{P}\cdot\text{R}}{\text{P}+\text{R}}.
  \label{eq:metrics}
\end{align}

Jaccard~\cite{jaccard1912distribution} is our \textbf{primary ranking metric} because it penalizes both false positives and false negatives, making it hard to gain score simply by expanding or shrinking the predicted set.
These set-matching metrics are standard in multi-label evaluation~\cite{zhang2014review}.
We also report the \textbf{exact-match rate}:
\begin{equation}
  \text{EM} = \frac{1}{N}\sum_{i=1}^{N} \mathbb{1}[P_i = G_i],
  \label{eq:exact_match}
\end{equation}
namely, the fraction of samples for which the predicted and ground-truth sets are \emph{identical} ($P = G$ as sets).

\paragraph{Over-Prediction Penalty.}
To separate calibrated predictions from indiscriminate guessing, we add a penalty factor:
\begin{align}
  \text{penalty} &= \min\!\left(1,\tfrac{|G|}{|P|}\right), \notag \\
  J_{\text{pen}} &= J \times \text{penalty}, \quad
  F_{1,\text{pen}} = F_1 \times \text{penalty}.
  \label{eq:penalty}
\end{align}
When $|P| \leq |G|$, no penalty is applied.
When a model predicts more HKL entries than appear in the ground truth, the penalty linearly reduces the score.
We use it as a complementary diagnostic rather than as the main ranking criterion.

\paragraph{Edge Cases.}
For degenerate inputs, metric behavior is defined as follows:
(1)~When $P = \emptyset$ and $G \neq \emptyset$, all metrics evaluate to 0---the model failed to produce any valid prediction.
(2)~When $G = \emptyset$ and $P \neq \emptyset$, all metrics also evaluate to 0.
(3)~When $P = G = \emptyset$, all metrics evaluate to 1.0 (vacuously correct).
In our benchmark, case~(3) does not occur as every sample has $|G| \geq 1$.

\paragraph{Aggregation.}
All metrics are computed per sample and aggregated via macro-averaging, so every sample carries equal weight.
We report results along four axes: overall, per dataset, per difficulty level (grouped by union size), and per $2\theta$ range (low/mid/high angle).

\section{Experiments}
\label{sec:experiments}

\subsection{Models}
\label{sec:models}

The evaluation covers seven frontier VLMs:
\textbf{Tier~1}: GPT-5.4 (OpenAI), Gemini~3.1~Pro (Google), Gemini~3.0~Pro (Google), Qwen3.6-Plus (Alibaba)\footnote{Qwen3.6-Plus is developed by Alibaba Cloud. The authors of this paper are affiliated with Alibaba Group. All models were evaluated under identical conditions using their public APIs.};
\textbf{Tier~2}: Doubao-Seed-2.0 (ByteDance);
\textbf{Tier~3}: Kimi-K2.5 (Moonshot AI), Claude~Opus~4.6 (Anthropic).
These tiers are descriptive groupings based on natural gaps in the Jaccard distribution (Figure~\ref{fig:overall}), not statistically derived thresholds.
All models were accessed through their respective APIs with thinking or reasoning mode enabled where available.

\paragraph{Inference Configuration.}
For Gemini~3.1~Pro and Gemini~3.0~Pro, we use \texttt{thinking\_budget=8192}; GPT-5.4 uses \texttt{reasoning\_effort=high}.
Qwen3.6-Plus, Doubao-Seed-2.0, and Kimi-K2.5 expose chain-of-thought reasoning by default in their APIs, and Claude~Opus~4.6 is queried in extended thinking mode.
In the reasoning or thinking modes used here, temperature and random seed are either fixed by the API (e.g., GPT-5.4 fixes temperature at 1 in reasoning mode) or not exposed as user-configurable parameters.
These configurations represent the strongest reasoning mode available for each model at the time of evaluation; the heterogeneity across providers is a limitation inherent to any cross-API benchmark.

If a model output could not be parsed into a valid \texttt{max\_peak\_hkls} list, we treated it as an empty prediction ($P = \emptyset$) and assigned zero to all metrics for that sample; no retry was performed.
Parse success rates vary across models, from 86\% (Claude~Opus~4.6, 35 failures) to 100\% (GPT-5.4, 0 failures), and are reported in Figure~\ref{fig:radar} (Appendix~\ref{app:suppfig}) under ``Parse Success Rate''.
All reported numbers come from a \emph{single run} per model; we discuss the implications of that choice in Section~\ref{sec:discussion}.

\subsection{Prompt Design}
\label{sec:prompt}

Each model receives three inputs: the XRD pattern image, the CIF crystal structure text, and the chemical formula.
The prompt asks the model to identify all Miller indices contributing to the highest-intensity peak and to return them as structured JSON in the form \texttt{\{"max\_peak\_hkls": [[h,k,l], ...]\}}.
The output convention is adapted to the crystal system so that models produce either standard 3-index HKLs or 4-index Miller-Bravais notation when required.
The full prompt template appears in Appendix~\ref{app:prompt}.

\section{Results}
\label{sec:results}

\subsection{Overall Performance}
\label{sec:overall}

\begin{figure}[t]
  \centering
  \includegraphics[width=0.95\linewidth]{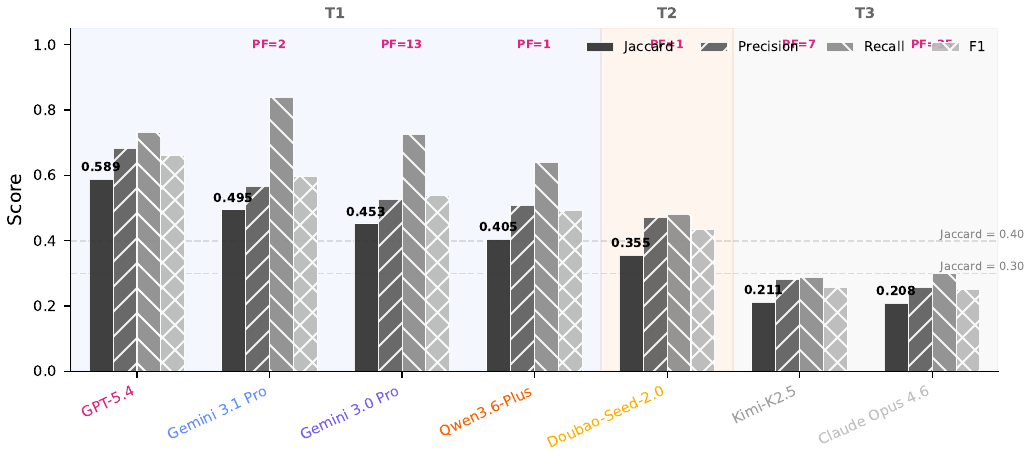}
  \caption{Overall performance across four base metrics. Horizontal dashed lines mark tier boundaries (T1: Jaccard $\geq 0.40$; T2: $0.30$--$0.40$). Within T1, GPT-5.4 leads on precision while the Gemini models are recall-dominant; T3 models remain below 0.30 on all metrics.}
  \label{fig:overall}
\end{figure}

Table~\ref{tab:overall} summarizes the aggregate performance of all seven models across the full benchmark. Models demonstrate considerable variation in capability, spanning nearly 0.38 Jaccard points from the strongest to the weakest result---a range that argues against treating frontier VLMs as interchangeable for this task. No model exceeds Jaccard~0.59, and six of seven sit well below 0.50 against a deterministic ceiling of 1.0, establishing that the benchmark remains far from solved. The distribution separates naturally into three tiers (Figure~\ref{fig:overall}): GPT-5.4 stands alone at the top, four models cluster in a mid-range band between Jaccard~0.40 and 0.50, and two trail well below 0.30.

\begin{table*}[t]
  \centering
  \caption{Overall performance of seven VLMs on \benchname{} (250 samples, macro-averaged). Models are grouped by tier (T1: Jaccard $\geq 0.40$; T2: $0.30$--$0.40$; T3: $< 0.30$). Bold indicates best per column. \textbf{Avg.\ $|\hat{S}|$}: mean predicted set size. \textbf{OP\%}: percentage of samples where $|\hat{S}| > |S^*|$ (over-prediction rate).}
  \label{tab:overall}
  \resizebox{\textwidth}{!}{%
  \begin{tabular}{cl cccc cc cc}
    \toprule
    & \textbf{Model} & \textbf{Jaccard}$\uparrow$ & \textbf{Prec.}$\uparrow$ & \textbf{Recall}$\uparrow$ & \textbf{F1}$\uparrow$ & $J_\text{pen}\uparrow$ & $F_{1,\text{pen}}\uparrow$ & \textbf{Avg.\ $|\hat{S}|$} & \textbf{OP\%} \\
    \midrule
    \textbf{T1} & GPT-5.4           & \textbf{0.5888} & \textbf{0.6828} & 0.7306 & \textbf{0.6597} & \textbf{0.5333} & \textbf{0.5837} & 2.61 & 27.6 \\
    & Gemini 3.1 Pro    & 0.4954          & 0.5653 & \textbf{0.8374} & 0.5974          & 0.3980          & 0.4553          & 4.99 & 53.2 \\
    & Gemini 3.0 Pro    & 0.4526          & 0.5254 & 0.7266          & 0.5376          & 0.3797          & 0.4293          & 4.53 & 44.8 \\
    & Qwen3.6-Plus      & 0.4049          & 0.5090 & 0.6386          & 0.4934          & 0.3398          & 0.3970          & 3.80 & 42.0 \\
    \midrule
    \textbf{T2} & Doubao-Seed-2.0   & 0.3549          & 0.4701 & 0.4788          & 0.4329          & 0.3140          & 0.3738          & 2.11 & 30.0 \\
    \midrule
    \textbf{T3} & Kimi-K2.5         & 0.2115          & 0.2807 & 0.2875          & 0.2568          & 0.1857          & 0.2196          & 2.43 & 34.0 \\
    & Claude Opus 4.6   & 0.2085          & 0.2564 & 0.2986          & 0.2513          & 0.1802          & 0.2104          & 2.36 & 31.2 \\
    \bottomrule
  \end{tabular}%
  }
\end{table*}

The primary axis of differentiation across these tiers is prediction strategy rather than raw capacity. GPT-5.4 achieves both the highest precision (0.6828) and a strong recall (0.7306), generating an average of only 2.61~HKLs per sample---essentially matching the ground-truth mean of 2.51---and over-predicting on just 27.6\% of cases. This compact-but-accurate strategy yields the highest Jaccard (0.5888) and exact-match rate (37.6\%) in the evaluation. The Gemini models pursue an orthogonal approach: Gemini~3.1~Pro holds the highest recall in the entire evaluation (0.8374), but achieves it by returning nearly five HKLs per sample on average, over-predicting on 53.2\% of cases, which depresses its precision to 0.5653---more than 0.12 points below GPT-5.4. Gemini~3.0~Pro follows a similar but slightly more conservative pattern (avg.\ 4.53~HKLs, 44.8\% over-prediction, Jaccard 0.4526). Qwen3.6-Plus occupies an intermediate position within T1 (Jaccard 0.4049; avg.\ 3.80~HKLs, 42\% over-prediction).

Doubao-Seed-2.0 stands apart in T2 with the most compact prediction set of any model (avg.\ 2.11~HKLs) and nearly balanced precision--recall (0.4701/0.4788), reflecting a conservative strategy that avoids over-prediction yet falls short at these absolute accuracy levels (Jaccard 0.3549). The T3 models---Kimi-K2.5 (Jaccard 0.2115, exact match 10.4\%) and Claude~Opus~4.6 (0.2085, exact match 10.0\%)---trail by a wide margin that cannot be attributed to set-size differences. Both return prediction sets of size comparable to GPT-5.4 (2.43 and 2.36~HKLs respectively), yet their accuracy is far lower, indicating that the bottleneck lies in identifying the correct HKLs rather than in calibrating how many to predict.

\subsection{Prediction Strategies and Over-Prediction}
\label{sec:strategy}

\begin{figure}[t]
  \centering
  \includegraphics[width=0.7\linewidth]{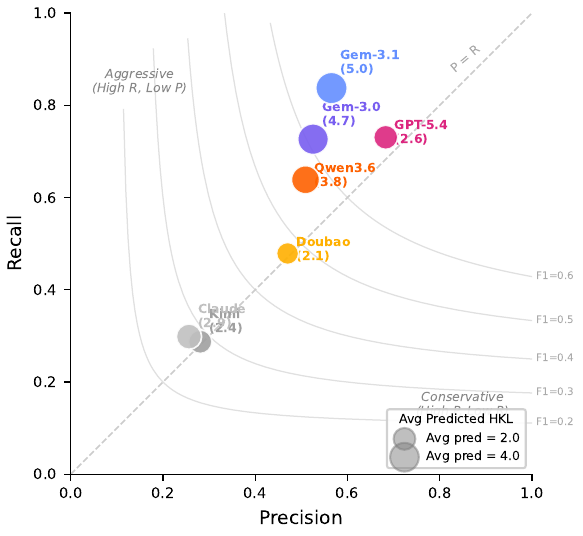}
  \caption{Precision-recall scatter with iso-F1 curves. Point size encodes average predicted set size. GPT-5.4 occupies the high-precision, high-recall quadrant; the Gemini models favor recall over precision; Doubao achieves the most balanced P/R profile.}
  \label{fig:pr_scatter}
\end{figure}

The precision--recall plane (Figure~\ref{fig:pr_scatter}) distills these strategy differences into a clear geometric separation. GPT-5.4 occupies the high-precision, high-recall quadrant with a recall-to-precision ratio of only 1.07---close to unity---reflecting a conservative but accurate prediction budget that stays near the ground-truth mean of 2.51~HKLs per sample. The Gemini models shift substantially toward recall, exhibiting R/P ratios of 1.38--1.48 and average prediction sizes of 4.5--5.0~HKLs per sample, roughly double the ground-truth mean. Qwen3.6-Plus is moderately aggressive (R/P ratio 1.25, avg.\ 3.8~HKLs), while Doubao-Seed-2.0 achieves the most balanced profile (R/P ratio 1.02, avg.\ 2.11~HKLs). Notably, the T3 models cluster in the low-precision, low-recall corner of the scatter despite R/P ratios near 1.0---balance achieved at low absolute values reflects an inability to identify correct HKLs rather than a well-calibrated search strategy.

\begin{figure}[t]
  \centering
  \includegraphics[width=0.85\linewidth]{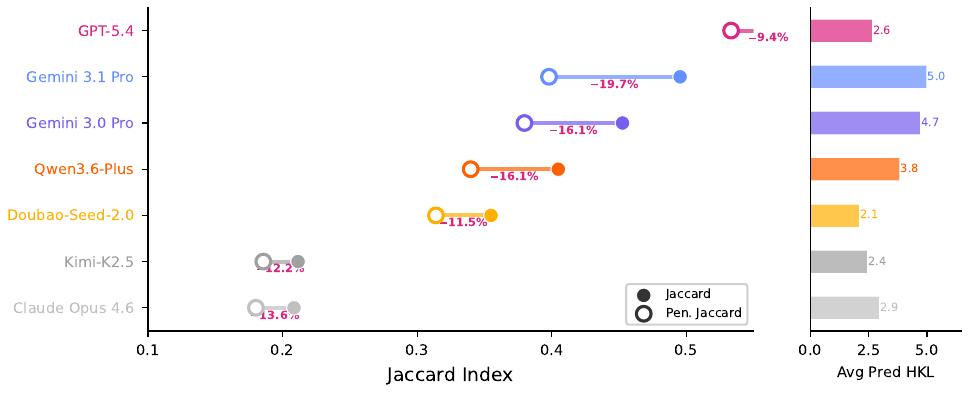}
  \caption{Effect of over-prediction penalty: dumbbell chart comparing Jaccard (left) and $J_\text{pen}$ (right) for each model. GPT-5.4 incurs the smallest penalty ($-9.4\%$) thanks to its compact prediction budget (avg.\ 2.61), while Gemini~3.1 suffers the largest ($-19.7\%$) due to aggressive predictions (avg.\ 5.0).}
  \label{fig:penalty}
\end{figure}

The over-prediction penalty makes these strategy differences directly measurable in performance terms. Gemini~3.1~Pro suffers the largest relative reduction ($-19.7\%$ on $J_\text{pen}$ relative to $J$), while GPT-5.4 loses only $-9.4\%$ (Figure~\ref{fig:penalty}). As a result, GPT-5.4's advantage over Gemini~3.1 grows from 0.093 Jaccard points to 0.135 in $J_\text{pen}$---an amplification with real practical significance: for downstream applications such as structure refinement, the difference between a 27.6\% and a 53.2\% over-prediction rate carries consequences that the raw 0.093 Jaccard gap somewhat understates.

\subsection{Per-Dataset Analysis}
\label{sec:perdataset}

\begin{figure}[t]
  \centering
  \includegraphics[width=0.85\linewidth]{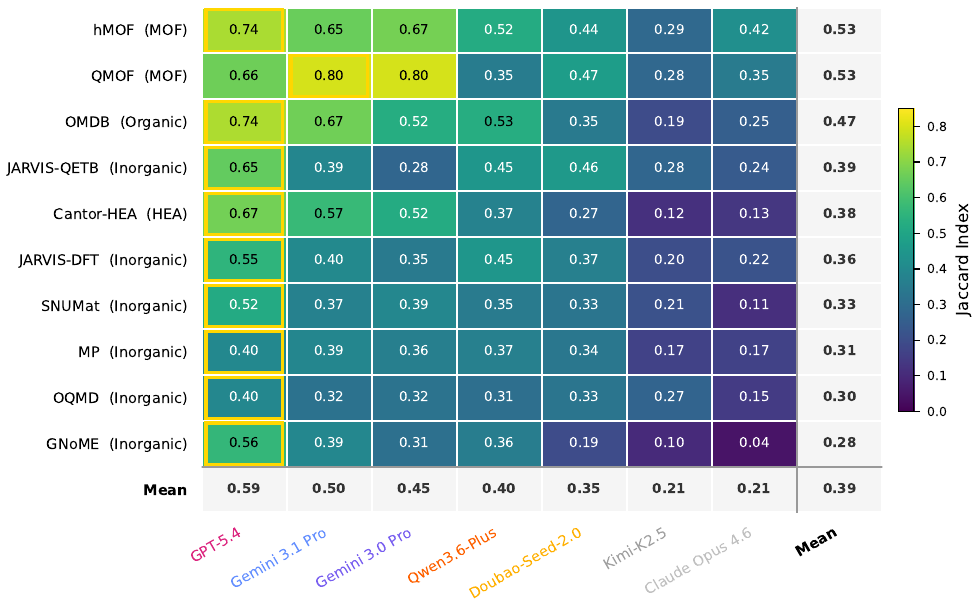}
  \caption{Per-dataset Jaccard heatmap (7 models $\times$ 10 databases). Gold borders mark each row's maximum. GPT-5.4 is best on 9 of 10 databases; only QMOF favors Gemini~3.1. The MOF datasets (QMOF, hMOF) tend to yield higher scores, while GNoME remains difficult across models.}
  \label{fig:heatmap}
\end{figure}

Figure~\ref{fig:heatmap} shows the per-dataset Jaccard heatmap. The tier ranking from Table~\ref{tab:overall} holds broadly across all databases, yet the magnitude of per-dataset variation is large enough to expose systematic patterns that aggregate scores conceal.

Performance clusters most naturally along the axis of structural complexity and angular crowding. The MOF and organic-crystal subsets---QMOF, hMOF, and OMDB---form a consistently high-performance group: GPT-5.4 achieves 0.741 on hMOF and 0.745 on OMDB, while QMOF yields the single highest cell value in the heatmap (Gemini~3.1~Pro: 0.796)---the only database on which GPT-5.4 does not lead (0.662 vs.\ 0.796). This elevated performance likely reflects structural properties of MOF and organic crystal patterns: their dominant reflections commonly appear at low $2\theta$ angles where peaks are well separated and overlap is minimal, reducing the difficulty of both visual extraction and HKL enumeration~\cite{rosen2021qmof}. In contrast, the AI-discovered inorganic compounds in GNoME constitute the hardest subset across the panel. GPT-5.4 scores only 0.563 on GNoME, while Claude~Opus~4.6 scores 0.043---a 13-fold disparity within the same 25 samples that likely reflects poor model generalization to compositions and structure types absent from standard training corpora.

The remaining inorganic databases (JARVIS-DFT, JARVIS-QETB, Cantor-HEA, SNUMat, OQMD, MP) occupy an intermediate band, with GPT-5.4 pushing several subsets above Jaccard~0.50 (GNoME: 0.563, JARVIS-QETB: 0.646, SNUMat: 0.517). Among mid-tier models, Qwen3.6-Plus shows relative strength on the JARVIS databases (JARVIS-DFT: 0.448; JARVIS-QETB: 0.446), as does Doubao-Seed-2.0 on JARVIS-QETB (0.457). Across all databases, GPT-5.4 leads on 9 of 10; each database contributes only 25 samples, so these per-database figures should be treated as directional rather than definitive.

\subsection{Difficulty and Angle-Range Analysis}
\label{sec:difficulty}

\begin{figure*}[t]
  \centering
  \includegraphics[width=0.95\linewidth]{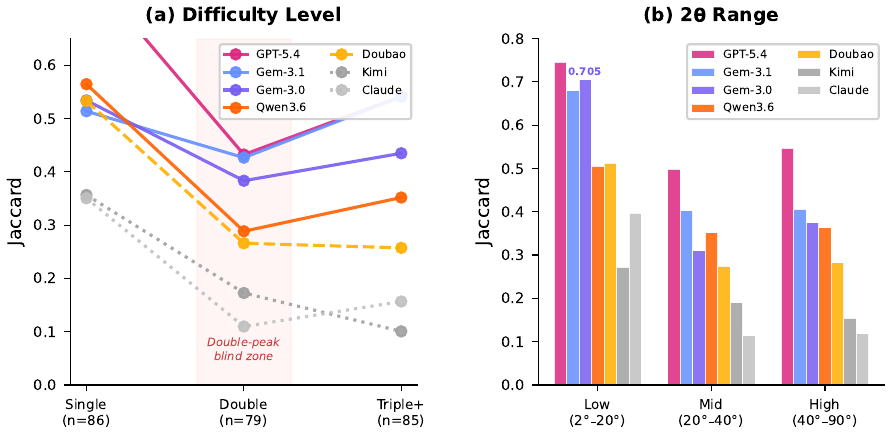}
  \caption{Performance decomposition: (a)~Jaccard vs.\ difficulty level (union size). GPT-5.4 leads all three levels; most models dip at union size 2 and recover partly at 3+. (b)~Jaccard by $2\theta$ range. Low-angle peaks ($2^\circ$--$20^\circ$) are easiest for nearly all models.}
  \label{fig:difficulty_theta}
\end{figure*}

Stratifying performance by union size and angular range reveals two further systematic patterns---one unexpected, one orderly---that together deepen the picture of where current VLMs struggle (Figure~\ref{fig:difficulty_theta}).

Across all evaluated models, performance as a function of union size follows a non-monotonic rather than uniformly declining curve, with the double-peak regime (union size~2) emerging as the most difficult setting for most models---harder than either single-peak or many-peak cases. GPT-5.4 leads at all three difficulty levels, reaching Jaccard~0.779 on single-peak samples (86 cases, 34.4\% of the benchmark)---more than 0.21 points above the next model, Qwen3.6-Plus (0.564). The aggressive recall strategy of Gemini~3.1, which benefits complex multi-peak patterns, becomes a liability here: its 0.514 on single-peak cases falls below both Gemini~3.0 (0.534) and Doubao-Seed-2.0 (0.533), models that predict fewer HKLs per sample. At union size~2, most models decline sharply: GPT-5.4 (0.432) and Gemini~3.1 (0.427) are nearly tied, while Qwen3.6-Plus collapses by 49\% relative to its single-peak score (0.564$\to$0.289). Partial recovery occurs at union size $\geq 3$, where GPT-5.4 (0.542) and Gemini~3.1 (0.541) co-lead while T3 models fall further behind (Kimi-K2.5: 0.101). The resulting non-monotonic difficulty curve is discussed in Section~\ref{sec:discussion}.

The angle-range breakdown follows a more orderly pattern. Low-angle peaks ($2^\circ$--$20^\circ$, 83 samples) are the easiest setting for every model: GPT-5.4 reaches Jaccard~0.746 in this range and Gemini~3.0 reaches 0.705---their strongest performances on any angle group. The advantage of the low-angle regime is twofold: reflections are more widely spaced, reducing visual extraction ambiguity, and MOF samples---which are structurally simpler for this task---are concentrated in this angular range (Table~\ref{tab:datasets}), further elevating scores. GPT-5.4 maintains its lead at mid angles (0.498) and high angles (0.546). The most dramatic angle dependence belongs to Claude~Opus~4.6, whose Jaccard drops by a factor of $3.5\times$ from 0.396 at low angles to 0.114 at mid angles---a sensitivity that none of the T1 models exhibit to the same degree, suggesting a disproportionate degradation in fine-grained visual extraction at intermediate angular density.

\subsection{Qualitative Examples}
\label{sec:examples}

\begin{figure*}[t]
  \centering
  \includegraphics[width=0.95\linewidth]{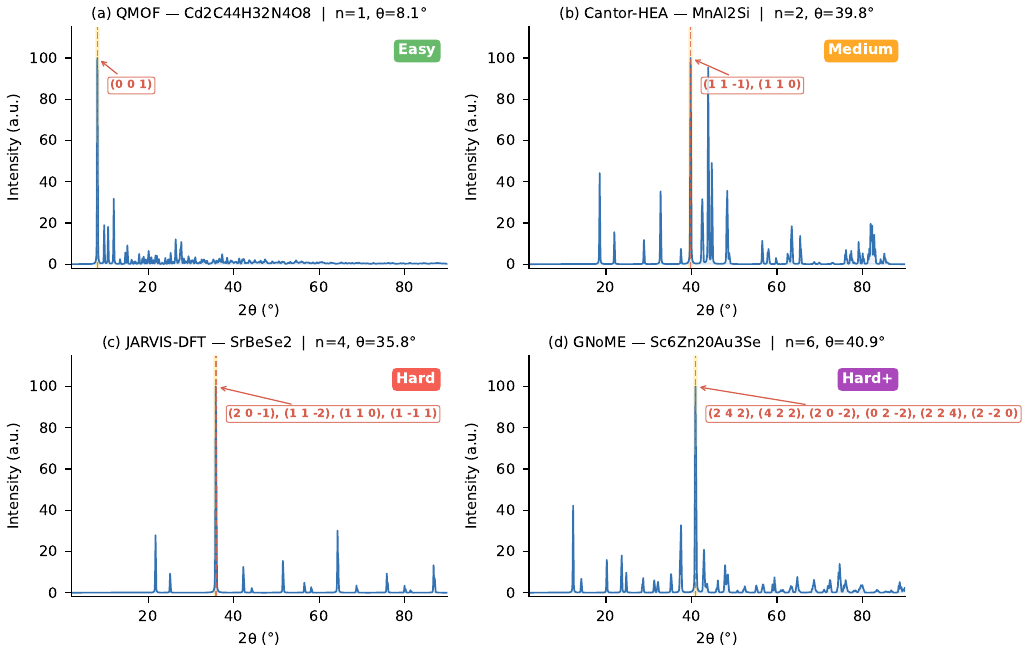}
  \caption{Representative XRD patterns from \benchname{}, ordered from easier to harder cases. \textbf{Easy}: WETVUN\_FSR (QMOF, Cd$_2$C$_{44}$H$_{32}$N$_4$O$_8$, union size 1). \textbf{Medium}: gra64680367 (Cantor-HEA, MnAl$_2$Si, union size 2), with two overlapping reflections. \textbf{Hard}: JVASP-74226 (JARVIS-DFT, SrBeSe$_2$, union size 4). \textbf{Hardest}: 958d370b0f (GNoME, Sc$_6$Zn$_{20}$Au$_3$Se, union size 6), with six contributing planes.}
  \label{fig:xrd_examples}
\end{figure*}

Figure~\ref{fig:xrd_examples} presents four representative cases spanning the full difficulty range, from a single-peak MOF pattern (union size~1) to a GNoME sample with six overlapping reflections (union size~6).

\section{Discussion and Limitations}
\label{sec:discussion}

A consistent observation across all seven evaluated models is the systematic decline of peak identification accuracy as target reflections shift toward higher $2\theta$ values. Gemini~3.0, for instance, drops from Jaccard~0.705 on low-angle samples to 0.375 above $40^\circ$---a near halving of performance---and the trend is not idiosyncratic to any single provider but is reproduced across the entire panel of models, mirroring documented difficulties of vision--language systems with closely spaced chart features~\cite{masry2022chartqa}. The mechanism behind this degradation is intuitive at the visual level: at higher angles, neighboring reflections subtend progressively smaller horizontal spans on the rendered plot, and the visual cues that distinguish one peak from its neighbour become correspondingly finer. Whether the bottleneck lies primarily in limited visual encoder resolution or in insufficient exposure to scientific plots during pretraining is something our data cannot disentangle; both pathways are plausible, and isolating their relative contribution would require controlled ablations beyond the scope of this benchmark.

A related but distinct challenge arises when we turn from perception to reasoning. Because the ground truth is deterministically derivable from each CIF via \texttt{pymatgen}'s \texttt{XRDCalculator}, a model that faithfully executed the underlying crystallographic computation could, in principle, achieve perfect scores from the CIF alone. The strongest model, GPT-5.4, nevertheless reaches only Jaccard~0.5888, leaving a gap of more than 0.41 to this deterministic ceiling. The gap is too large to attribute to visual extraction noise alone and instead points to a structural limitation: current VLMs do not appear to translate CIF input into an executable calculation chain (cell parameters $\rightarrow$ $d$-spacings $\rightarrow$ $2\theta$ positions $\rightarrow$ HKL labels), but rather pattern-match against fragments of crystallographic knowledge. Taken together with the angle-range trend, these results suggest that two complementary deficits---fine-grained visual discrimination and structured numerical reasoning---compound across the task, so that improvements along only one axis are unlikely to close the observed gap on their own.

Beyond model capability, the structure of our benchmark itself reveals a dataset characteristic that resists simple difficulty narratives. The union-size-2 regime is unexpectedly the hardest setting: five of seven models perform worse on double-peak cases than on either single-peak or many-peak cases, a pattern incompatible with any monotonic difficulty model in which more contributing reflections imply a harder problem. We attribute this non-monotonicity to the awkward middle ground occupied by two-peak overlaps---they remove the unambiguous cue of a lone contributor that single-peak cases provide, yet they do not yet offer the structural redundancy that sometimes rewards aggressive multi-HKL prediction on larger unions. Confirming this mechanism will require controlling for confounds such as angular distribution and crystal-system composition within each union-size group, an analysis we leave to future work.

This non-monotonicity, in turn, illustrates the broader diagnostic value of the benchmark. Because each sample bundles the XRD image, the CIF text, and the chemical formula, \benchname{} can disentangle failure modes in ways that an image-only benchmark cannot: poor performance despite CIF access implicates reasoning failures, performance decay across angle ranges implicates visual extraction, and elevated over-prediction rates---27.6\% for GPT-5.4 versus 53.2\% for Gemini~3.1~Pro---reveal when a model compensates for uncertainty by inflating its candidate set rather than abstaining. For the broader community, these orthogonal signals offer concrete handles on model development: visual decoders can be evaluated and improved in isolation through angle-range stratified scoring, reasoning chains can be probed by holding the image fixed and manipulating CIF availability, and calibration can be tracked through over-prediction rates without conflating it with raw accuracy. We see this disentangling capacity, rather than any single aggregate score, as the most useful contribution of \benchname{} to ongoing efforts in scientific multimodal evaluation.

Several scope decisions qualify how broadly these conclusions transfer. All XRD patterns in \benchname{} are computationally generated from CIF data and therefore lack the noise, baseline drift, preferred orientation, amorphous background contributions, and thermal broadening that characterise laboratory diffractograms; experimental powder diffractometers typically exhibit FWHM in the range $0.08^\circ$--$0.30^\circ$ for well-crystallised samples, with broader peaks for nanocrystalline or poorly crystallised materials, and signal-to-noise ratios vary widely with instrument and sample preparation. Model performance on experimental data is therefore likely to be lower than the figures we report, although the relative ordering of models---driven as it is by reasoning rather than purely instrumental factors---should be more robust to this shift. A second scope choice concerns crystallographic equivalence: HKL comparison uses strict tuple matching, so $[1\,0\,0]$ and $[\bar{1}\,0\,0]$ count as distinct predictions, and the 38 cubic samples in our data---15\% of the total, each carrying up to 48 equivalent reflections---receive no credit for symmetry-equivalent answers. While this design choice introduces a conservative bias against high-symmetry systems, it preserves a single unambiguous scoring rule across all crystal systems; symmetry-aware evaluation via point-group folding is a natural and high-priority extension that we expect to narrow the reported gaps without altering their qualitative ordering.

A further limitation is the absence of a human expert baseline. Without such a reference, it is genuinely difficult to judge whether a Jaccard of 0.5888 represents a narrow or wide gap relative to expert performance on the same task. In related domains such as chart question answering, human baselines typically range from 0.80 to 0.95~\cite{masry2022chartqa}, but XRD HKL identification from rendered plots may in fact be harder for unaided experts than standard chart reading, since it demands explicit crystallographic calculation beyond visual inspection; the human ceiling on our task is therefore an open empirical question rather than something that can be borrowed from neighbouring benchmarks. Two additional caveats further bound the present results: all models were evaluated with a single prompt template (Appendix~\ref{app:prompt}), so prompt variations such as chain-of-thought scaffolding or few-shot examples could shift the observed failure modes, and the evaluation granularity itself is uniform---a $\pm 0.30^\circ$ tolerance applied across the full $2\theta$ range, 25 samples per database within a 250-sample total, and a single inference run per model whose reasoning modes either fix temperature internally or do not expose it as a user-configurable parameter. These choices keep the present comparisons internally consistent but mean that fine-grained distinctions, such as the T1/T2 boundary between Qwen3.6-Plus (0.4049) and Doubao-Seed-2.0 (0.3549) at $\Delta{=}0.050$, should be interpreted cautiously. Establishing a human ceiling, broadening the prompt distribution, and quantifying multi-run variance together form the most immediate methodological agenda we plan to pursue in subsequent extensions of \benchname{}.

\section{Conclusion}
\label{sec:conclusion}

\benchname{} tests a specific scientific capability that existing multimodal benchmarks do not cover: reading a diffraction curve at sub-degree precision and mapping its features to crystallographic labels.
The evaluation of seven frontier VLMs reveals three failure modes that would remain invisible in coarser benchmarks: angle-dependent visual degradation, shallow crystallographic reasoning despite access to CIF text, and a non-monotonic difficulty profile driven by double-peak overlap.
Even GPT-5.4, the strongest model (Jaccard~0.5888), falls more than 0.41 below the deterministic ceiling that a simple CIF-based calculation achieves (a \texttt{pymatgen}-based calculation that achieves Jaccard~1.0 by construction)---evidence that end-to-end neural inference alone is insufficient for this task under current architectures.
Coupling a VLM front-end with a deterministic crystallographic solver is one promising path forward.

Because each sample pairs the XRD image with CIF text and chemical formula, the benchmark can attribute failures to specific bottlenecks---visual extraction, crystallographic reasoning, or calibration under overlap---rather than collapsing them into a single score.
The most pressing extensions are symmetry-aware scoring via point-group folding and modality ablations (image-only, CIF-only, full-input) benchmarked against a deterministic CIF-based upper bound.
Evaluating on experimental XRD patterns from open repositories such as COD~\cite{grazulis2009cod}, stratified by noise level and crystallinity, would further test how well these conclusions transfer beyond synthetic data.

\section*{Data Availability}
The complete benchmark dataset, including all 1{,}936 XRD pattern images and ground-truth annotations, is publicly available at:
\url{https://huggingface.co/datasets/xiaodu-ali/CrystalXRD-Bench}.


\bibliographystyle{unsrtnat}
\bibliography{references}

\appendix

\section{VLM Prompt Template}
\label{app:prompt}

Below is the prompt template sent to each VLM. The model receives the XRD image as a visual input alongside this text:

\begin{quote}
\small
\begin{verbatim}
You are a materials science expert specializing
in X-ray diffraction (XRD) analysis. You are given:
1. An XRD pattern image (intensity vs. 2theta)
2. The CIF of the material: <CIF_TEXT>
3. The chemical formula: <FORMULA>

Task: Identify ALL Miller indices (hkl) for the
HIGHEST peak. It may result from superposition of
multiple crystallographic planes.
Return JSON:
{"max_peak_hkls": [[h,k,l], ...]}
\end{verbatim}
\end{quote}

For hexagonal and trigonal systems, the prompt switches to 4-index Miller-Bravais notation $[hkil]$ with the constraint $i = -(h+k)$.

\section{Detailed Per-Dataset Results}
\label{app:detailed}

Table~\ref{tab:detailed_jaccard} reports the full per-dataset Jaccard scores, and Table~\ref{tab:detailed_penjaccard} reports the corresponding penalized Jaccard scores.

\begin{table*}[h]
  \centering
  \caption{Per-dataset Jaccard scores (7 models $\times$ 10 databases). Bold indicates the best model for each database.}
  \label{tab:detailed_jaccard}
  \resizebox{\textwidth}{!}{%
  \begin{tabular}{l ccccccc}
    \toprule
    \textbf{Database} & \textbf{GPT-5.4} & \textbf{Gemini 3.1} & \textbf{Gemini 3.0} & \textbf{Qwen3.6} & \textbf{Doubao} & \textbf{Kimi} & \textbf{Claude} \\
    \midrule
    Cantor-HEA  & \textbf{0.670} & 0.573 & 0.520 & 0.372 & 0.273 & 0.117 & 0.133 \\
    GNoME       & \textbf{0.563} & 0.392 & 0.309 & 0.356 & 0.194 & 0.103 & 0.043 \\
    hMOF        & \textbf{0.741} & 0.655 & 0.673 & 0.520 & 0.444 & 0.288 & 0.419 \\
    JARVIS-DFT  & \textbf{0.553} & 0.399 & 0.346 & 0.448 & 0.368 & 0.200 & 0.218 \\
    JARVIS-QETB & \textbf{0.646} & 0.391 & 0.278 & 0.446 & 0.457 & 0.278 & 0.242 \\
    MP          & \textbf{0.397} & 0.391 & 0.363 & 0.367 & 0.337 & 0.169 & 0.167 \\
    OMDB        & \textbf{0.745} & 0.666 & 0.525 & 0.530 & 0.347 & 0.193 & 0.253 \\
    OQMD        & \textbf{0.395} & 0.320 & 0.325 & 0.307 & 0.330 & 0.271 & 0.147 \\
    QMOF        & 0.662 & \textbf{0.796} & 0.795 & 0.353 & 0.469 & 0.283 & 0.349 \\
    SNUMat      & \textbf{0.517} & 0.371 & 0.393 & 0.352 & 0.331 & 0.213 & 0.113 \\
    \bottomrule
  \end{tabular}%
  }
\end{table*}

\begin{table*}[h]
  \centering
  \caption{Per-dataset Penalized Jaccard ($J_\text{pen}$) scores. Bold indicates the best model for each database.}
  \label{tab:detailed_penjaccard}
  \resizebox{\textwidth}{!}{%
  \begin{tabular}{l ccccccc}
    \toprule
    \textbf{Database} & \textbf{GPT-5.4} & \textbf{Gemini 3.1} & \textbf{Gemini 3.0} & \textbf{Qwen3.6} & \textbf{Doubao} & \textbf{Kimi} & \textbf{Claude} \\
    \midrule
    Cantor-HEA  & \textbf{0.645} & 0.488 & 0.421 & 0.303 & 0.237 & 0.102 & 0.096 \\
    GNoME       & \textbf{0.518} & 0.282 & 0.227 & 0.295 & 0.169 & 0.090 & 0.027 \\
    hMOF        & \textbf{0.668} & 0.564 & 0.615 & 0.446 & 0.415 & 0.266 & 0.390 \\
    JARVIS-DFT  & \textbf{0.482} & 0.275 & 0.257 & 0.386 & 0.281 & 0.157 & 0.165 \\
    JARVIS-QETB & \textbf{0.564} & 0.259 & 0.194 & 0.353 & 0.377 & 0.204 & 0.198 \\
    MP          & \textbf{0.328} & 0.301 & 0.287 & 0.315 & 0.307 & 0.128 & 0.148 \\
    OMDB        & \textbf{0.705} & 0.584 & 0.468 & 0.502 & 0.337 & 0.193 & 0.234 \\
    OQMD        & \textbf{0.329} & 0.208 & 0.241 & 0.224 & 0.291 & 0.240 & 0.138 \\
    QMOF        & 0.616 & 0.735 & \textbf{0.751} & 0.280 & 0.460 & 0.264 & 0.327 \\
    SNUMat      & \textbf{0.478} & 0.284 & 0.336 & 0.294 & 0.267 & 0.213 & 0.079 \\
    \bottomrule
  \end{tabular}%
  }
\end{table*}

\section{Supplementary Figures}
\label{app:suppfig}

\begin{figure}[h]
  \centering
  \includegraphics[width=0.9\linewidth]{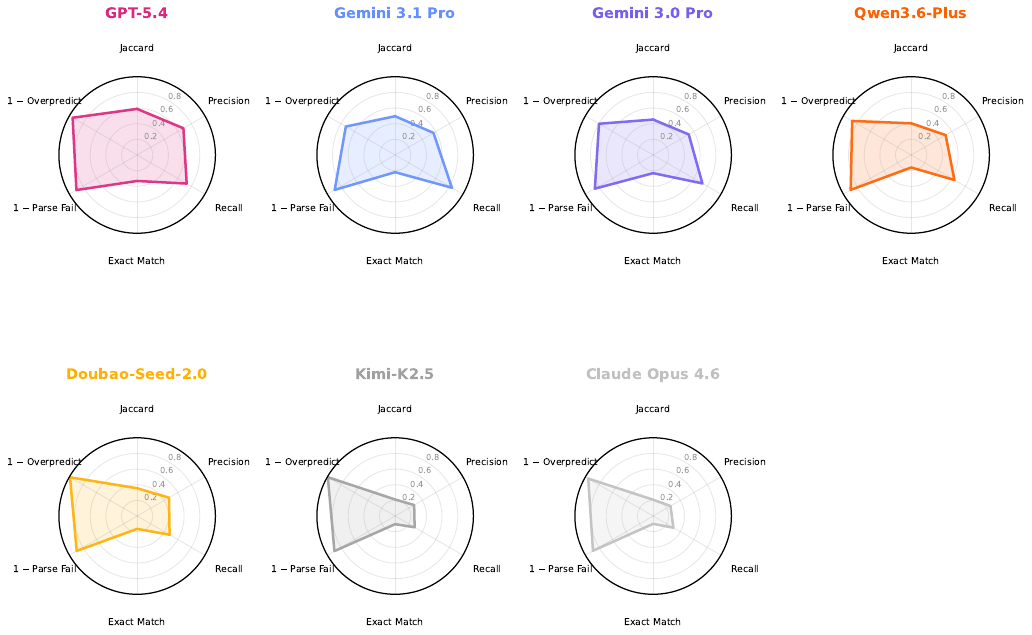}
  \caption{Radar charts for all models over six dimensions: Jaccard, Precision, Recall, Exact Match Rate, Parse Success Rate, and Conservativeness (the inverse of over-prediction rate).}
  \label{fig:radar}
\end{figure}

\begin{figure}[h]
  \centering
  \includegraphics[width=0.9\linewidth]{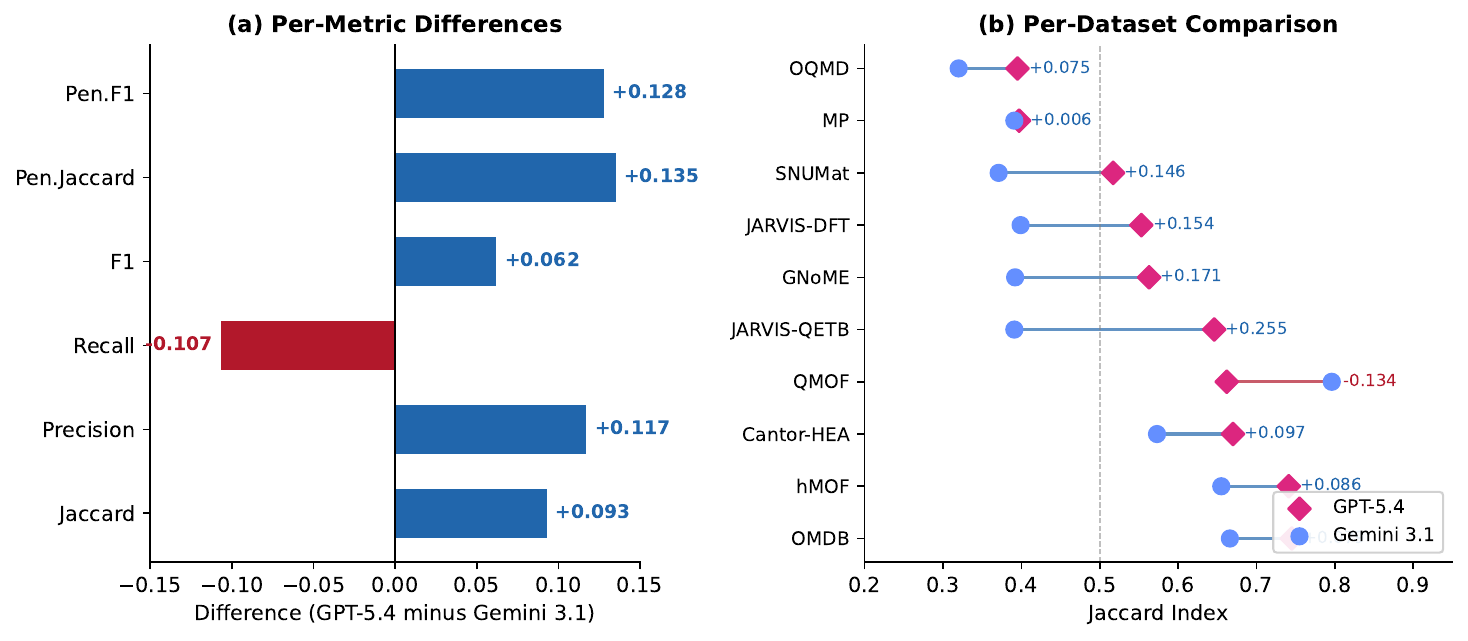}
  \caption{Top-two model comparison (GPT-5.4 vs.\ Gemini~3.1~Pro): (a)~waterfall chart of per-metric differences, and (b)~per-dataset paired comparison. GPT-5.4 leads in Jaccard (+0.093), Precision (+0.118), and exact-match rate (37.6\% vs.\ 22.0\%), while Gemini~3.1 retains higher Recall (0.837 vs.\ 0.731) through a more aggressive prediction strategy.}
  \label{fig:gemini_compare}
\end{figure}

\section{KMeans Sampling Validation}
\label{app:silhouette}

Table~\ref{tab:silhouette} reports silhouette scores for the KMeans clustering ($k{=}25$) used to select samples from each database.
Silhouette scores range from $-1$ to $+1$, with higher values indicating better-defined clusters.
The values here are moderate (mean 0.228), consistent with the heterogeneity of the feature space, where materials within a single database vary widely in crystal system, space group, and peak configuration.
In our setting, KMeans is used mainly to distribute samples across the feature space rather than to recover sharply separated natural clusters, so moderate scores are acceptable for the sampling objective.

\begin{table}[h]
  \centering
  \caption{KMeans clustering silhouette scores ($k{=}25$) for each database. Features: peak position ($\theta^*$), peak intensity, number of theoretical peaks, union size, crystal system, and space group number.}
  \label{tab:silhouette}
  \begin{tabular}{lcc}
    \toprule
    \textbf{Database} & \textbf{Pool Size} & \textbf{Silhouette Score} \\
    \midrule
    QMOF        & 199 & 0.201 \\
    SNUMat      & 188 & 0.203 \\
    GNoME       & 183 & 0.212 \\
    MP          & 199 & 0.214 \\
    hMOF        & 199 & 0.221 \\
    OMDB        & 201 & 0.222 \\
    Cantor-HEA  & 195 & 0.223 \\
    OQMD        & 200 & 0.240 \\
    JARVIS-DFT  & 172 & 0.245 \\
    JARVIS-QETB & 200 & 0.297 \\
    \midrule
    \textbf{Mean} & \textbf{1,936} & \textbf{0.228} \\
    \bottomrule
  \end{tabular}
\end{table}

\end{document}